\documentclass[a4paper,11pt]{article}
\usepackage[T1]{fontenc}
\usepackage{amsfonts}
\usepackage{verbatim}
\usepackage{amsmath, amsthm, amssymb, mathrsfs}
\usepackage{graphicx}
\usepackage[style=numeric-comp]{biblatex}
\usepackage{placeins}
\usepackage{caption}
\usepackage{subcaption}

\newtheorem{definition}{Definition}
\newtheorem{theorem}{Theorem}
       
\begin{document}

\begin{titlepage}
   \begin{center}
       \vspace*{1cm}

       {\Large \textbf{Stochastic Scaling in Loss Functions for \\ Physics-Informed Neural Networks}}
            
       \vspace{5cm}

By\\
       {Ethan Mills \\ Alexey Pozdnyakov} 

       \vspace{1.5cm}
            
       In partial fulfillment of a MRC-REU Program Research Project
under the supervision of Dr.\ Duan Chen

       \vspace{0.8cm}

       Summer 2022\\
       UNC Charlotte\\
            
\vspace{5.5cm}

   \end{center}
\end{titlepage}

\section*{Abstract}
Differential equations are used in a wide variety of disciplines, describing the complex behavior of the physical world. Analytic solutions to these equations are often difficult to solve for, limiting our current ability to solve complex differential equations and necessitating sophisticated numerical methods to approximate solutions. Trained neural networks act as universal function approximators, able to numerically solve differential equations in a novel way. In this work, methods and applications of neural network algorithms for numerically solving differential equations are explored, with an emphasis on varying loss functions and biological applications. Variations on traditional loss function and training parameters show promise in making neural network-aided solutions more efficient, allowing for the investigation of more complex equations governing biological principles.

\section{Introduction}
We begin by introducing the topic of differential equations. The study of these equations is broad, as these equations qualitatively relate unknown functions with their derivatives. Real world phenomena that involve dynamic systems and rates of change can be described by these equation, necessitating algorithms for solving such equations in fields such as physics, biology, and chemistry. Both ordinary differential equations (ODE), involving unknown functions of one independent variables, and partial differential equations (PDE), involving more than one independent variable, are useful tools for modeling real life processes. Phenomena such as nerve impulse transmission, chemical reactions, growth of tumors, and molecular dynamics are just a few examples \cite{jones_2010}.

Some simple differential equations can be solved analytically. However, due to the complex nature of many differential equations, analytic solutions are not always possible. Numerical methods for solving have been developed and can aid in approximating solutions. Methods such as Euler's Method and the Runge-Kutta 4th order method are sufficient for solving some ODEs \cite{stoer_bulirsch_2011}. More sophisticated methods such as finite difference methods and finite element methods are required for solving PDEs \cite{hoffman_frankel_2021}. Numerical methods are limited by numerous factors, including equation complexity, often leading to increased computational expense while solving. Higher dimensional solutions are also costly, further complicated by functions with numerical instability. New methods for solving differential equations numerically are needed, as many real life problems are too computationally expensive to compute with classical numerical methods. In this paper, we introduce the idea of neural networks as a tool for solving differential equations numerically, and we propose a way to improve upon current methods.

Artificial neural networks consist of one or more interconnected layers of neurons, with an input layer, many hidden layers, and an output layer. The output of each neuron is determined by the input, weights, and activation function \cite{wang_2003}. Neural networks are able to effectively approximate functions using a learning algorithm which utilizes backpropagation and stochastic gradient descent \cite{elbrachter_perekrestenko_grohs_bolcskei_2021}.
Furthermore, given enough neurons, they are able to approximate any continuous function to an arbitrary degree of accuracy, resulting from the universal approximation theorem \cite{Cybenko1989}\cite{hornik_stinchcombe_white_1989}. Neural networks are also simple to compute, with the output of a neuron $h_i$ given by
\begin{equation}
    h_i = \rho \left(\sum_{j=1}^NA_{ij}x_j + b_i\right),
\end{equation}
where $\rho : \mathbb{R} \to \mathbb{R}$ is the activation function, $N$ is the number of input neurons for this layer, $A_{ij}$ are the weights, $x_j$ the inputs, and $b_i$ the bias \cite{wang_2003}.

Neural networks are able to deal more efficiently with higher dimensional inputs, making them more favorable than other numerical methods that rely on mesh methods, which is computationally expensive in higher dimensions. Furthermore, neural networks are able to work with noisy data as they solve an optimization problem of best fit when approximating functions. Finally, neural networks are able to work with complicated domains and surfaces, a limitation of classical numerical methods that rely on mesh and grid problem solving.

Due to the novel nature of neural networks and machine learning, current methods of numerically solving with them faces some challenges, which we aim to address and improve upon in this paper. The question of optimal network size is an unanswered one \cite{elbrachter_perekrestenko_grohs_bolcskei_2021}, limiting optimal network design and increasing run time and error. Moreover, the training process of neural networks is not fully understood, and thus, vanishing gradients and complex loss-space presents issues while training.

In this paper, we aim to provide a brief introduction to neural networks, improve current algorithms, and provide a biological application. In section 2, relevant neural network algorithms and theory will be introduced, as well as physics informed neural networks (PINNS) - the main method by which neural networks are used to solve differential equations. In section 3, we introduce loss function methodology and look at variations of the standard static loss functions. In the following section, experimental results with different loss functions are provided and analyzed. In section 5, we apply these methods to a biologically-relevant example, the nonlinear Poisson-Boltzmann equation. Finally, in section 6, we discuss relevant results and future work.

\section{Neural Network Background}
    \subsection{Neural Networks and Loss functions}
\begin{figure}[t]
    \begin{center}
         \includegraphics[scale=0.52]{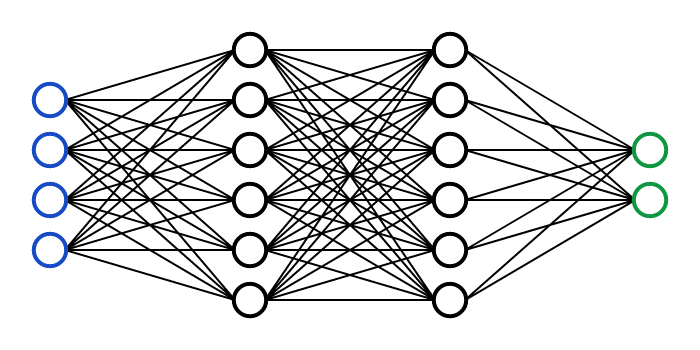}
    \end{center}
    \caption{\sf FCNN with $L=3$, and $(N_0, N_1, N_2, N_3) = (4, 6, 6, 2)$. Note that each layer corresponds to the activation of an affine transformation, denoted $\rho \circ W_\ell$ in Definition \ref{d-NN}. Moreover, each contiguous pair of layers is a fully connected bipartite graph.}
    \label{f-NN}
    \end{figure}

    We begin this section with a mathematical description of fully connected neural networks (FCNN). Note that there are other neural network models such as convolution neural networks (CNN), recurrent neural networks (RNN), etc. We focus on FCNN since these are the standard model for physics informed neural networks (PINN), the main of object of study in this paper. 
    \begin{definition}
        Let $L \in \mathbb{N}$ and $N_0, \ldots, N_L \in \mathbb{N}$. Then a neural network $\phi$ with activation function $\rho : \mathbb{R} \to \mathbb{R}$ is a map $\phi:\mathbb{R}^{N_0} \to \mathbb{R}^{N_L}$ defined by 
        \begin{equation}
            \phi(x) = \begin{cases}
            W_1,  & L = 1 \\
            W_2 \circ \rho \circ W_1, & L =2 \\
            W_L \circ \rho \circ W_{L-1} \circ \rho \circ \ldots \circ \rho \circ W_1, & L \geq 3.
        \end{cases}
        \end{equation}
        where
        \begin{equation}
            W_\ell(x) = A_\ell x + b_\ell \text{ with } A_\ell \in \mathbb{R}^{N_\ell \times N_{\ell -1}} \text{ and } b \in \mathbb{R}^{N_\ell} \text{ } \forall \ell \in \{1,\ldots, L \}.
        \end{equation}
        Note that $\rho$ acts component wise on vectors.
    \label{d-NN}
    \end{definition}
    Following the conventions of \cite{https://doi.org/10.48550/arxiv.1901.02220}, we call $L$ the depth of the network, and $\max \{N_0, \ldots, N_L\}$ the width of the network. We also call the elements of each $A_\ell$ the weights of the network and the elements of each $b_\ell$ the biases of the network. All together, the weights and biases form the parameters of the neural network, and we denote this collection of values by $\theta$. We denote the space of all possible choices of parameters by $\Theta$. The term fully connected simply refers to that fact that we do not enforce any elements in $A_\ell$ or $b_\ell$ to be $0$. This has a natural interpretation in the graph representation of a neural network, seen in Figure $\ref{f-NN}$. Moreover, we will focus on neural networks with $\rho = \tanh$ and $N_1 = N_2 = \ldots = N_{L-1}$, since these are common in the PINN literature. Next, we describe the way in which the parameters of the neural network are determined.
    
    All neural network models are trained on a dataset of paired inputs and outputs. Moreover, they are trained to minimize a loss function $\mathcal{L} : \Theta \to \mathbb{R}_{\geq 0}$. Note that the loss function depends on the dataset, but this is usually fixed beforehand. Thus we consider the optimization problem 
    \begin{equation}
        \min_{\theta \in \Theta} \mathcal{L}(\theta).
    \label{p-min}
    \end{equation}
    Depending on the specific loss function, this problem can vary greatly in dimensionality, convexity, and computational complexity. We will delve deeply into loss functions in section 3, but for now we just consider it as a differentiable function mapping each $\theta \in \Theta$ to a non-negative real number.
    \subsection{Backpropagation and Automatic Differentiation}
    
    In order to solve Problem \ref{p-min}, we use variations of gradient descent. This requires a method for taking gradients of the form 
    \begin{equation}
        \frac{\partial \mathcal{L}}{\partial W_\ell} \text{ and } \frac{\partial \mathcal{L}}{\partial b_\ell} \text{ } \forall \ell \in \{1, \ldots, L\},
    \end{equation}
    which is inefficient to do directly with neural networks of high depth or high width. Instead, we take these gradients using the well known backpropagation algorithm. This algorithm computes these gradients inductively using the chain rule. Let us define $a_\ell$ and $z_\ell$ to be the values at each layer after and before activation, meaning $a_\ell = \rho(W_\ell a_{\ell-1} + b_\ell)$ and $z_\ell = W_\ell a_{\ell-1}+b_\ell$. Next consider the quantities
    \begin{align}
        \delta_L &= \frac{\partial \mathcal{L}}{\partial W_L} \odot \rho'(z_L) \label{e-deltaL} \\
        \delta_{\ell} &= ((W_{\ell+1})^T\delta_{\ell+1}) \odot \rho'(z_{\ell}), \text{ } \forall \ell \in \{1, \ldots,L-1\}, \label{e-deltal}
    \end{align}
    where $\odot$ is the Hadamard product. Note that these values can be efficiently computed via dynamic programming. Moreover, $\forall \ell \in \{1,\ldots,L-1\}$,
    \begin{equation}
        \frac{\partial \mathcal{L}}{\partial b_\ell^j} = \delta_\ell^j \text{ and } \frac{\partial \mathcal{L}}{\partial W_\ell^{ij}} = a_{\ell-1}^j \delta_\ell^i \label{e-grads}
    \end{equation}
    where superscripts correspond to the index. Using Equations \ref{e-deltaL},\ref{e-deltal}, and \ref{e-grads}, we now have an efficient way to compute gradients of the loss function with respect to the neural network parameters.
    
    It is also important that we can compute gradients of the form $\nabla_x\phi$ where $x \in \mathbb{R}^{N_0}$ is the input to the neural network. This is key to computing the loss functions involved in PINNs, and this is done using an algorithm called automatic differentiation, or autodiff. This is a generalization of the backpropagation algorithm that computes the chain rule inductively for any function that can be represented as an acyclic computational graph, such as a FCNN. The idea is very similar to backpropagation, so we will not give a separate mathematical description. For a thorough review of autodiff, see \cite{JMLR:v18:17-468}.
    
    \subsection{Gradient Descent and ADAM}
    
    Through backpropagation we can effectively compute the gradient of $\mathcal{L}$, and thus we can perform gradient descent. In the most basic form, we can describe this algorithm by
    \begin{equation}
        \theta_{n+1} = \theta_n - \eta \nabla \mathcal{L}(\theta)
    \end{equation}
    where $\eta$ is a carefully chosen learning rate, and $n \in \{1, \ldots, N_{steps}\}$. There is a large amount of research that has gone into improving this algorithm for various different optimization problems. In the case of neural networks, we have to optimize this algorithm for handling high dimensional loss functions and for computational efficiency. This includes variations such as batch gradient descent or stochastic gradient descent, which compute the gradient based on subsets of the parameters in the former, and which add random noise to the gradient in the latter. 
    
    The current standard for training neural networks is the ADAM algorithm, first proposed in \cite{https://doi.org/10.48550/arxiv.1412.6980}. This method is designed to choose step sizes based on the first and second moments of the gradients. The authors performed extensive experiments that show ADAM outperforms the other state of the art gradient descent algorithms in neural network training. In our experiments, we use the default parameters set by TensorFlow ($\beta_1 = 0.9, \beta_2 = 0.999, \epsilon = 1e-7$) except for the learning rate $\alpha$. For the learning rate, we test two different schedules - piecewise constant decay, and cyclical. In piecewise constant decay, the learning rate is defined by a piecewise constant function of the training step. We test this method since it is common in the PINN literature. We also test a cyclical learning rate, since this was shown in \cite{https://doi.org/10.48550/arxiv.1506.01186} to not only remove the need to fine tune the learning rate, but also to improve the final accuracy of the neural network model.
    
    \subsection{Universal Approximation Theorem}
    
    We now give a key theoretical result about neural networks. Namely, it has been proven by Cybenko in \cite{Cybenko1989} that a FCNN can approximate any real valued continuous function on a compact domain.
    \begin{theorem}[Universal Approximation Theorem with Arbitrary width]
        Let $C(X,Y)$ denote the set of all continuous functions $f : X \to Y$. Let $\rho \in C(\mathbb{R},\mathbb{R})$, and let $\rho \circ x$ denote $\rho$ applied to each component of $x$. Then $\rho$ is not polynomial iff for every $n,m \in \mathbb{N}$, compact $K \subset \mathbb{R}^n$, $f \in C(K,\mathbb{R}^m)$,$\epsilon > 0$, there exists $k \in \mathbb{N}$, $A \in \mathbb{R}^{k \times n}$, $b \in \mathbb{R}^k$, and $C \in \mathbb{R}^{m \times k}$ such that
        \label{t-UAT}
    \begin{equation}
        \sup_{x \in K} ||f(x)-g(x)|| < \epsilon \text{ where } g(x) = C \cdot (\rho \circ (A \cdot x + b))
    \end{equation}
    Note : similar theorems exist which fix the width but require arbitrary depth 
    \end{theorem}
    With this theorem, we know that there exists a single layer neural network with $\rho = \tanh$ which approximates any $f \in C(K,\mathbb{R}^m)$. Moreover, we can extend this to any function which has a finite number of discontinuities. This is done using the discontinuity capturing shallow neural network (DCSNN) technique described in \cite{https://doi.org/10.48550/arxiv.2106.05587}. The key idea is that we can extend a $d$-dimensional piecewise continuous function to a $(d+1)$-dimensional continuous function by adding a new variable to parameterize the domain. For simplicity, consider the the case where $f:\Omega \to \mathbb{R}^m$ is piecewise continuous and given by
    \begin{equation}
    f(x) = \begin{cases}
        f^+(x) \text{ if $x \in \Omega^+$} \\
        f^-(x) \text{ if $x \in \Omega^-$}
        \end{cases},
        \label{e-DCSNN1}
    \end{equation}
    where $\Omega = \Omega^+ \sqcup \Omega^-$. We can then define $f_{\text{aug}} : \Omega \times [-1,1] \to \mathbb{R}$ given by
    \begin{equation}
        f_{\text{aug}}(x) = \frac{1+z}{2} f^+(x) + \frac{1-z}{2} f^-(x),
        \label{e-DCSNN2}
    \end{equation}
    which is continuous, and which allows us to recover $f^+$ and $f^{-}$ by evaluating it at $z=1$ and $z=-1$ respectively. This will be used in Section 5 to deal with a PDE which contains discontinuous coefficient functions.
    
    \subsection{Physics Informed Neural Networks (PINNs)}
    
    We now describe our primary tool for solving differential equations - PINNs. This technique was first introduced by Raissi et al. in \cite{https://doi.org/10.48550/arxiv.1711.10561}. This technique takes advantage of the fact that neural networks can approximate the solutions of differential equations by the universal approximation theorem, as well as the fact that differential equations can be encoded into the loss function using the autodiff algorithm.
    
    Following the notation of \cite{https://doi.org/10.13140/rg.2.2.20057.24169}, consider the following general parameterised PDE problem:
    \begin{align}
    \begin{split}
        &\text{PDE : }\mathcal{F}(\hat{\mathbf{u}}, \frac{\partial \hat{\mathbf{u}}}{\partial t}, \frac{\partial \hat{\mathbf{u}}}{\partial \mathbf{x}}, \frac{\partial^2\hat{\mathbf{u}}}{\partial t^2}, \frac{\partial^2 \hat{\mathbf{u}}}{\partial \mathbf{x}^2}, \ldots ; \mu) = 0 \text{,  } \mathbf{x} \in \Omega \subset \mathbb{R}^n, t\in \Upsilon \subset \mathbb{R} \\
        &\text{B.C. : }
        \mathcal{B}_i(\hat{\mathbf{u}}, \frac{\partial \hat{\mathbf{u}}}{\partial \mathbf{x}}, \frac{\partial^2 \hat{\mathbf{u}}}{\partial \mathbf{x}^2}, \ldots) = 0 \text{,  } \mathbf{x} \in \Gamma_i \text{, } i \in \{1, \ldots,N_{bc}\} \\
        &\text{I.C. : }
        \mathcal{I}(\hat{\mathbf{u}}, \frac{\partial \hat{\mathbf{u}}}{\partial t}, \frac{\partial^2 \hat{\mathbf{u}}}{\partial t^2}, \ldots) = 0 \text{,  } t \in \Upsilon_i \text{, } i \in \{1, \ldots,N_{ic}\}
    \end{split}
    \end{align}
    
\begin{figure}[h]
    \begin{center}
         \includegraphics[scale=0.259]{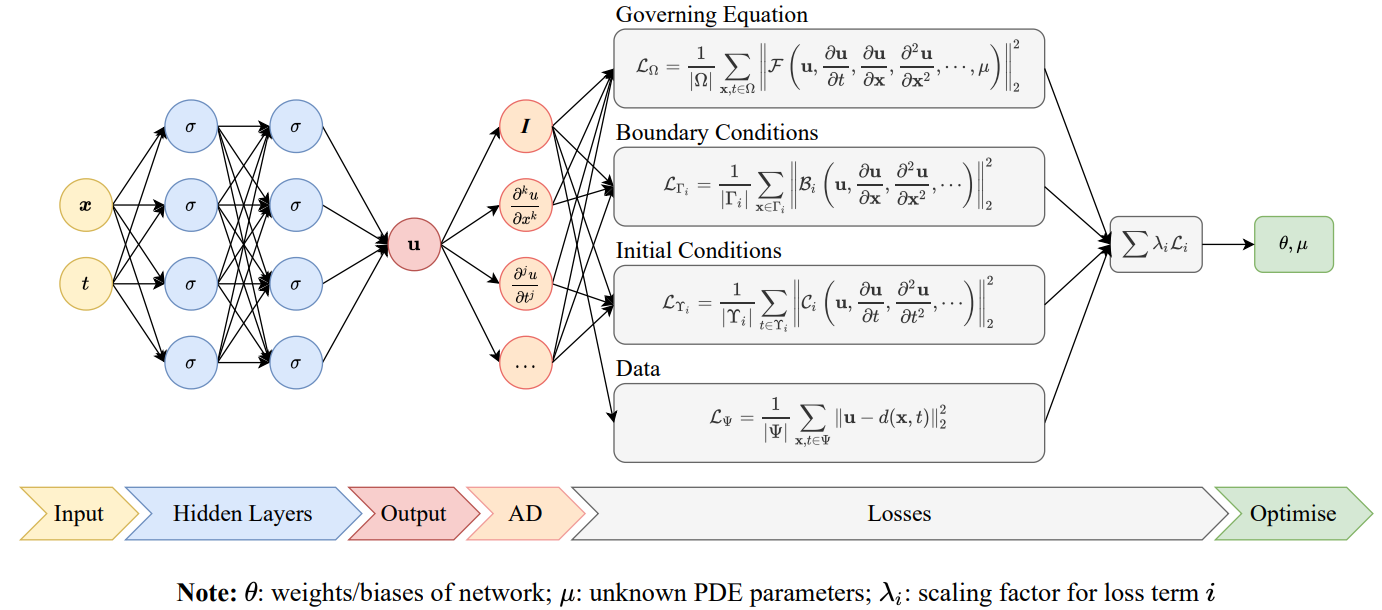}
    \end{center}
    \caption{\sf Schematic summary of the PINN model provided in \cite{https://doi.org/10.13140/rg.2.2.20057.24169}}
    \label{f-PINN}
\end{figure}

    We can find a neural network $\mathbf{u}$ which satisfies these equations by adding these terms into the loss function. In particular, suppose we take a finite random sample of points $\hat{\Omega} \subset \Omega \times \Upsilon$, $\hat{\Gamma}_i \subset \Gamma_i \text{ } \forall i \in \{1,\ldots,N_{bc}\}$, and $\hat{\Upsilon}_i \subset \Upsilon_i \text{ }\forall i \in \{1,\ldots,N_{ic}\}$. We may also have some set of points $\hat{\Psi}$ corresponding to known values of $\mathbf{u}$ at various points in the domain, denoted by $d(\mathbf{x},t)$. This is unnecessary for a well-posed PDE problem, but can aid in training and can often be obtained from lab data for PDEs arising from real world applications. Given these sets of points, we can introduce the following terms into our loss function:
    \begin{align}
    \begin{split}
            \mathcal{L}_{\Omega} &= \frac{1}{|\hat{\Omega}|} \sum_{\mathbf{x},t \in \hat{\Omega}} \Big\lVert \mathcal{F}(\mathbf{u}, \frac{\partial \mathbf{u}}{\partial t}, \frac{\partial \mathbf{u}}{\partial \mathbf{x}}, \frac{\partial^2\mathbf{u}}{\partial t^2}, \frac{\partial^2 \mathbf{u}}{\partial \mathbf{x}^2}, \ldots ; \mu) \Big\lVert^2, \\
            \mathcal{L}_{\Gamma_i} &= \frac{1}{|\hat{\Gamma}_i|}\sum_{\mathbf{x} \in \hat{\Gamma}_i} \Big\lVert \mathcal{B}_i(\mathbf{u}, \frac{\partial \mathbf{u}}{\partial \mathbf{x}}, \frac{\partial^2 \mathbf{u}}{\partial \mathbf{x}^2}, \ldots) \Big\lVert^2 \text{, } i \in \{1,\ldots, N_{bc}\}, \\
            \mathcal{L}_{\Upsilon_i} &= \frac{1}{|\hat{\Upsilon}_i|}\sum_{\mathbf{x} \in \hat{\Upsilon}_i} \Big\lVert \mathcal{I}_i(\mathbf{u}, \frac{\partial \mathbf{u}}{\partial t}, \frac{\partial^2 \mathbf{u}}{\partial t^2}, \ldots) \Big\lVert^2 \text{, } i \in \{1,\ldots, N_{ic}\},\\
            \mathcal{L}_\Psi &= \frac{1}{|\hat{\Psi}|} \sum_{\mathbf{x}, t \in \hat{\Psi}} \lVert \mathbf{u} - d(\mathbf{x},t) \lVert^2,
    \label{e-Lt}
    \end{split}
    \end{align}
    for some choice of norm $\lVert \cdot \lVert$. Finally, if we take our loss function to be
    \begin{equation}
        \mathcal{L}(\theta) = \mathcal{L}_\Omega(\theta) + \sum_{i=1}^{N_{bc}} \mathcal{L}_{\Gamma_i}(\theta) + \sum_{i=1}^{N_{ic}} \mathcal{L}_{\Upsilon_i}(\theta) + \mathcal{L}_\Psi(\theta),
        \label{e-sL}
    \end{equation}
    we have a neural network which approaches the solution of the PDE as the loss function approaches 0. While using the loss function in Equation \ref{e-sL} is standard, we can also consider various linear combinations of the terms defined in Equation \ref{e-Lt}. In Section 3, we will cover state of the art adaptive algorithms for choosing such linear combinations, as well as some novel stochastic algorithms. We conclude this section with a schematic summary of the PINN model (Figure \ref{f-PINN}).
    
\section{Loss Function Methodology}
    \subsection{Loss Balance and Multi-Objective Optimization}
    The question of selecting $\lambda_i$'s to optimize the performance of PINNs has been addressed in \cite{https://doi.org/10.13140/rg.2.2.20057.24169}, and we will briefly summarize the state of the art techniques. First, we give the following definitions from multi-objective optimization theory.
    \begin{definition}
         A solution $\hat{\theta} \in \Theta$ is said to Pareto dominate a solution $\theta \in \Theta$ (denoted $\hat{\theta} \prec \theta$) iff $\forall i, \mathcal{L}_i(\hat{\theta}) \leq \mathcal{L}_i(\theta)$, and $\exists j$ such that $\mathcal{L}_j(\hat{\theta}) < \mathcal{L}_j(\theta)$. Moreover, a solution $\hat{\theta}$ is said to be Pareto optimal if $\forall \theta \in \Theta, \hat{\theta} \prec \theta$.
    \end{definition}
    
    One important result is that the Pareto optimal solution to a multi-objective optimization problem is independent of the scalerization of the problem \cite{Ruchte2021EfficientMO}. Thus, we are free to update our scalerization as we wish, and we will not change the Pareto optimal solution as a result. We can take advantage of this fact to scale the loss terms with the goal of balancing the different terms, as well as keeping the terms normalized. This is particularly important in problems where we may see very large gradients coming from the boundary condition or the initial condition loss terms.
    \subsection{Adaptive algorithms}
    
    The state of the art algorithm presented in \cite{https://doi.org/10.13140/rg.2.2.20057.24169} is called Relative Loss Balancing with Random Lookback (ReLoBRaLo). This algorithm combines two other loss balancing algorithms and introduces a random lookback. The first of these algorithms is called Learning Rate Annealing (LRA), and is described by the following equations. At optimization step $t$, we scale the $k$ non-PDE loss terms according to 
    \begin{align}
        &\hat{\lambda}_i(t) = \frac{\max\{\nabla_\theta \mathcal{L}_\Omega(t)\}}{|\overline{\nabla_\theta \mathcal{L}_{\{\Gamma, \Upsilon\}}(t)}|}, \qquad i \in \{1,\ldots,k\}, \label{e-lra1} \\
        &\lambda_i(t) = \alpha\lambda_i(t-1)+(1-\alpha)\hat{\lambda}_i(t), \qquad i \in \{1,\ldots,k\}, \label{e-lra2}
    \end{align}
    where $\overline{\nabla_\theta \mathcal{L}_{\Gamma_i}(t)}$ is the mean of the gradients of $\mathcal{L}_{\Gamma_i}$ w.r.t. $\theta$, and $\alpha \approx 0.9$. Equation \ref{e-lra1} is meant to reduce the loss coming from B.C. or I.C. terms with large average gradients relative to the gradients coming from PDE the term, helping balance the loss terms. Equation \ref{e-lra2} smooths the balancing and helps avoid drastic changes between optimization steps. This same method is used to smooth the balancing in ReLoBraLo.
    
    The other algorithm incorporated in ReLoBRaLo is called SoftAdapt, and it is given by the following equation. At optimization step $t$, we scale all $k$ loss terms according to
    \begin{equation}
        \lambda_i(t) = \frac{\exp\big(\frac{\mathcal{L}_i(t)}{\mathcal{L}_i(t-1)}\big)}{\sum_{j=1}^k \exp \big( \frac{\mathcal{L}_j(t)}{\mathcal{L}_j(t-1)}\big)}, \qquad i \in \{1,\ldots,k\},
    \end{equation}
    where $\mathcal{L}_i(t)$ represents the loss from the $i$-th term at optimization step $t$. This algorithm balances the loss terms based on only the previous loss, giving a more computational efficient update rule. It also uses softmax to normalize the loss terms. We now present the ReLoBRaLo algorithm, which is described by
    \begin{align}
        &\hat{\lambda}_i^{(t;t')} = m \cdot \frac{\exp\Big(\frac{\mathcal{L}_i^{(t)}}{\mathcal{T}\mathcal{L}_i^{(t')}}\Big)}{\sum_{j=1}^k \exp \Big( \frac{\mathcal{L}_j^{(t)}}{\mathcal{T}\mathcal{L}_j^{(t')}}\Big)}, \qquad i \in \{1,\ldots,k\}, \label{e-RLB1}\\
        &\lambda_i^{(t)} = \alpha(\rho \lambda_i^{(t-1)} + (1-\rho)\hat{\lambda}_i^{(t;0)}) + (1-\alpha)\hat{\lambda}_i^{(t;t-1)}, \qquad i \in \{1,\ldots,k\} \label{e-RLB2},
    \end{align}
    where $\rho$ is a Bernoulli random variable with $\mathbb{E}(\rho)$ chosen to be close to 1, $\mathcal{T} \in (0, \infty)$, $\alpha \in (0, 1)$, and $m$ is the number of I.C. terms. Note that Equation $\ref{e-RLB1}$ is just a generalized SoftAdapt that allows us to update the values based on a loss term from time $t' < t$, as well as to control the hyperparameter $\mathcal{T}$. In particular, as $\mathcal{T} \to 0$, Equation $\ref{e-RLB1}$ approaches an argmax function, meanwhile $\mathcal{T} \to \infty$ causes Equation $\ref{e-RLB1}$ to return a uniform scaling. Moreover, Equation $\ref{e-RLB2}$ generalizes Equation $\ref{e-lra2}$ by allowing it to randomly choose between $\lambda_i^{(t-1)}$ and $\hat{\lambda}_i^{(t;0)}$ as the first term. This algorithm is extensively tested in \cite{https://doi.org/10.13140/rg.2.2.20057.24169}, and was shown to reach better testing accuracy than other algorithms with certain PDE problems. In particular, it outperformed other loss balance algorithms by an entire order of magnitude when tested on the Kirchhoff Plate Bending Equation.
    
    \subsection{Stochastic combinations}
    
    In this section we present stochastic combinations of loss functions, inspired by the use of a random variable in ReLoBRaLo. To the best of our knowledge, this is a novel approach to training neural networks, particularly in the context of PINNs. In general, we consider loss functions of the form
    \begin{equation}
        \mathcal{L}(\theta) = \sum_{i=1}^k \Lambda_i \mathcal{L}_i(\theta),
    \end{equation}
    where each $\Lambda_i$ is a strictly positive random variable. In section 4, we experiment with normally distributed random variables of the form $\Lambda_i \sim \text{Norm}(\mu_i(t), \sigma_i(t))$ where $\mu_i(t)$ and $\sigma_i(t)$ are the mean and standard deviation associated for the $i$-th loss term at optimization step $t$. We also test uniformly distributed random variables of the form $\Lambda_i \sim \text{Uni}(L_i(t), U_i(t))$ where $L_i(t)$ and $U_i(t)$ are the lower and upper bounds associated for the $i$-th loss term at optimization step $t$.
    
    We believe that having a loss space which is dynamic in this way could help the training process traverse complex loss functions. In particular, the authors of  \cite{https://doi.org/10.13140/rg.2.2.20057.24169} claim that the random lookback allows the model to escape from local minima in the loss space. We believe that having a stochastic loss space could help the model not only escape local minima, but also to keep the loss terms balanced. If the training process over-minimizes one term, it is likely that a future optimization step will scale that term down, causing it to step toward minima in the other loss terms. In section 4, we provide some evidence that this method can outperform the standard fixed loss function, as well as ReLoBRaLo, in certain PDE problems. 

\section{Results}
    In this section we will provide experimental results in which we test loss functions with various norms and with various stochastic scalings. We test the norms on two PDE problems - Burgers' Equation (\ref{e-burg}) and the Riccati Equation (\ref{e-ricc}). In addition to these two problems, we test the stochastic loss functions on the Poisson-Boltzmann Equation (\ref{e-PB}) as a scientific application of our work. We provide a thorough introduction to the Poisson-Boltzmann equation in Section 5, as well as its application to molecular biology and drug design. 
    
    For Burgers' equation, we consider the following problem. Let $\nu = 0.01$, $L=8$, $T=5$, and consider
    \begin{align}
    \begin{split}
        &\frac{\partial u}{\partial t} + u \frac{\partial u}{\partial x} - \nu \frac{\partial^2 u}{\partial x^2} = 0 \text{ for } x,t \in[0,L] \times [0,T] \\
        &u(x,0) = \exp{\big(-(x-L/2)^2\big)} \text{ for } x \in [0,L]\\
        &u(0,t) = u(L,t) = 0 \text{ for } t\in[0,T] \label{e-burg}.
    \end{split}    
    \end{align}
    Note that this equation has no analytic solution, so we compare our results to a numerical solution obtained using Scipy \cite{2020SciPy-NMeth}.

    For the Riccati equation, let $T = 0.99$ and consider
    \begin{equation}
        \frac{d^2 y}{d t^2} + 2t\bigg(\frac{d y}{d t}\bigg)^2 = 0 \text{ for } t\in[0,T]\text{, } y(0) = 2\text{, } y'(0)=-1. \label{e-ricc}
    \end{equation}
    In this case, we do have an analytic solution given by
    \begin{equation}
        y(t) = \frac{1}{2}(\ln|t-1| - \ln|t+1|) + 2.
    \end{equation}
    
    We test Burgers' Equation since this is a very standard problem in the PINN literature, and we test the Riccati Equation since it is singular at $t=1$. In particular, it is difficult for a neural network to model near $t=1$, so this problem provides us with a simple yet challenging to model equation. Additionally, we test the Poisson-Boltzmann equation since it has multiple interface conditions, as well as a very complicated interface geometry when used to model proteins - hopefully resulting in a complex loss space. Note that we do not have a numerical solution to the non-linear version of this equation to compare our solution to. Instead, we compare the performance of the stochastic loss functions based on a standard loss, which we define as $\mathcal{L}(\theta) = \sum_{i=1}^k \mathcal{L}_i(\theta)$. We do compare our solution to a numerical solution of the linearized equation in section 5.4. Finally, note that we compare all stochastic loss functions to the performance of a regular loss function (Reg), as well as to the performance of ReLoBRaLo (Rlb).
    
    \subsection{Optimal Norm}
\begin{figure}[h]
    \begin{center}
         \includegraphics[scale=0.5]{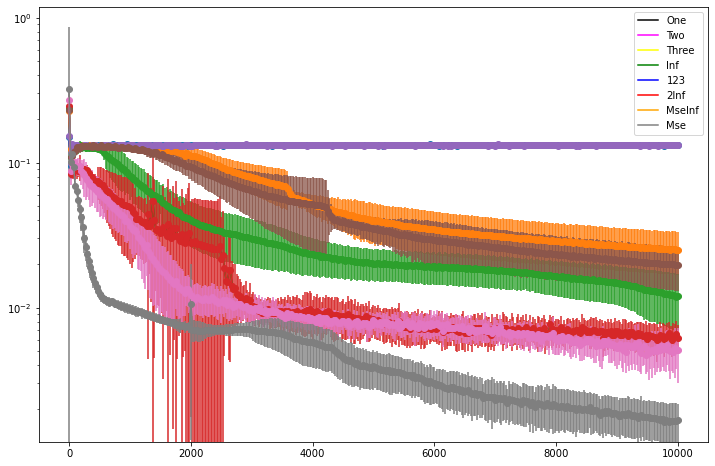}
         \includegraphics[scale=0.5]{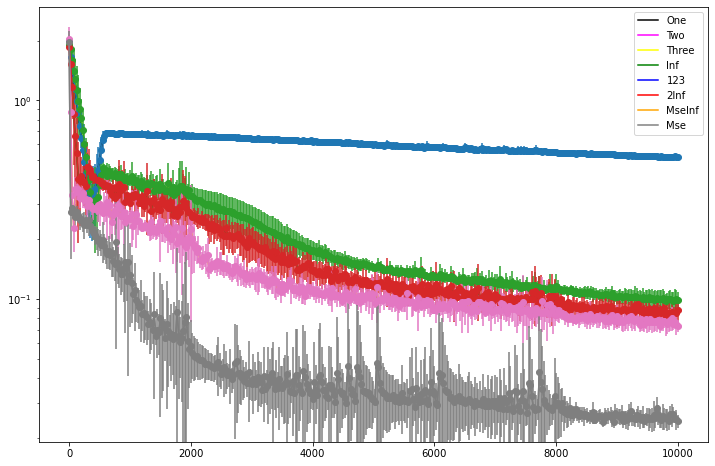}
    \end{center}
    \caption{\sf Average test MSE throughout 10000 epochs and over 10 trials for Burgers Equation (top) and the Riccati Equation (bottom). Error bars correspond to one standard deviation. Note that some norms led to diverging loss functions, so they have been omitted from the plot.}
    \label{f-norms}
\end{figure}

    Before testing the stochastic loss functions, we would like to provide evidence for the best choice of norm in Equation $\ref{e-Lt}$. The standard in PINNs and function approximation with neural networks is to use mean square error (MSE). Our experiments confirm that this norm outperforms other standard norms ($L^1, L^2, L^3, L^\infty$), as well as some sums of norms ($L^1 + L^2+L^3, L^2 + L^\infty, \text{MSE}+L^\infty$). In particular, we plot the average MSE between the neural network and the reference solution over 10 training trials for both Burgers' Equation and the Riccati Equation (Figure \ref{f-norms}). We call this value the test MSE, and it is the standard metric for model performance in the PINN literature. We also see that using MSE for the norm leads to significantly lower test MSE than any other norm. Note that test MSE is different from the loss, which is a measure of the error in the PDE, B.C., and I.C. of the neural network. Because of this, we believe the experiment is not biased toward MSE.

    \subsection{Fixed Variance Random Coefficients}
    Next, we will consider basic stochastic loss functions in which the random variable coefficients remain constant. In particular, we consider
    \begin{align}
    \begin{split}
        &\Lambda_i \sim \text{Norm}(1, 0.25) \qquad i \in \{1,\ldots,k\}, \text{ and }\\ 
        &\Lambda_i \sim \text{Uni}(0.5, 1.5) \qquad i \in \{1,\ldots,k\}.
        \label{eq-FVRC}
    \end{split}
    \end{align}
    The resulting average test MSE using these random variables with a piecewise constant learning rate in given in Figure \ref{f-PLR}. With a cyclic learning rate, the results are seen in Figure \ref{f-CLR}. We also test these loss function on the Poisson-Boltzmann Equation in Figure \ref{f-lysL}.

    \subsection{Decreasing Variance Random Coefficients}
\begin{figure}[h]
    \begin{center}
         \includegraphics[scale=0.5]{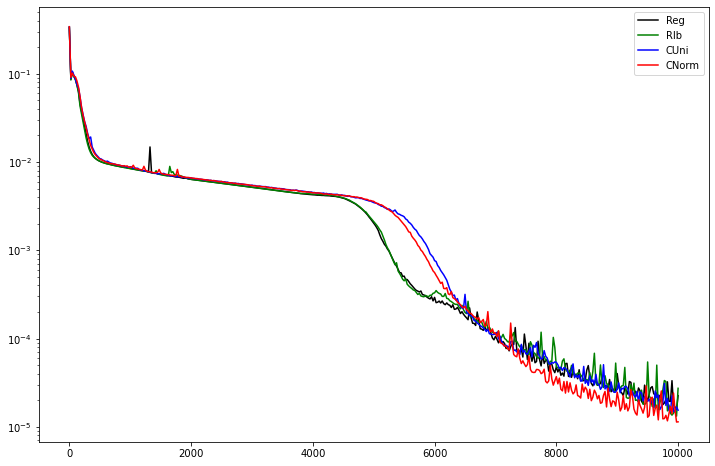}
         \includegraphics[scale=0.5]{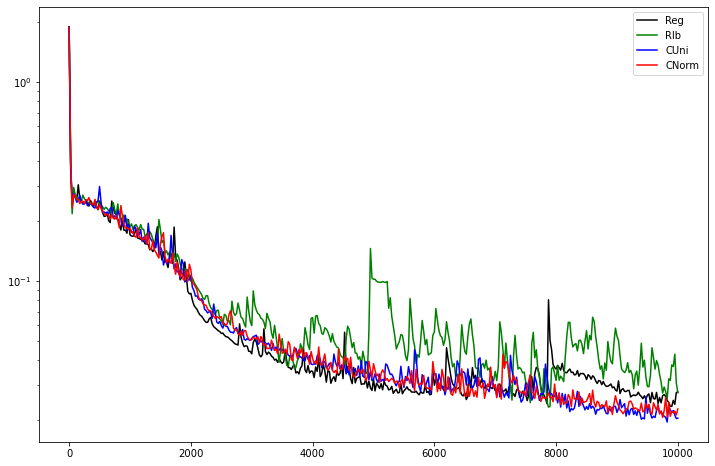}
    \end{center}
    \caption{\sf Average test MSE throughout 10000 epochs and over 10 trials for Burgers Equation (top) and the Riccati Equation (bottom). The learning rate begins at $10^{-2}$ and decays to  $2\cdot10^{-3}$ and  $5\cdot10^{-4}$ after epoch $2000$ and $8000$ respectively.}
    \label{f-PLRc}
\end{figure}

\begin{figure}[h]
    \begin{center}
         \includegraphics[scale=0.5]{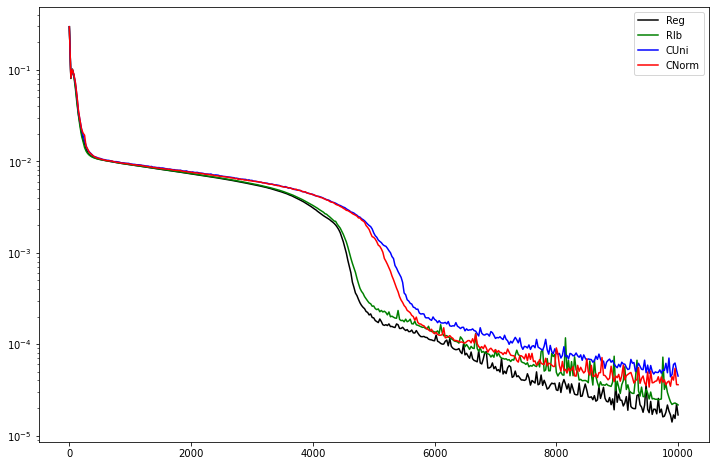}
         \includegraphics[scale=0.5]{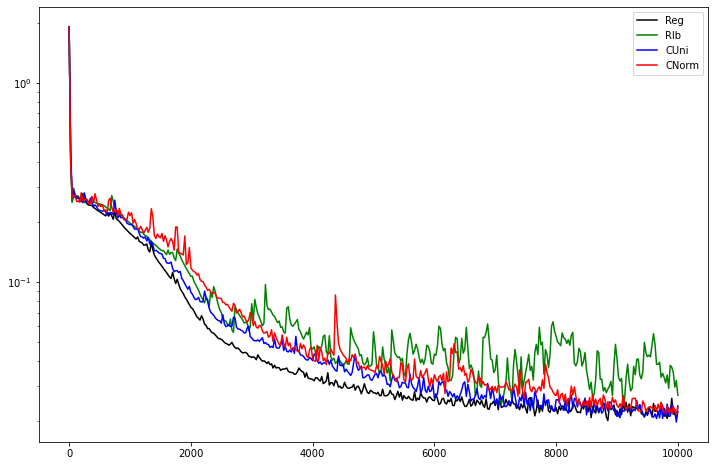}
    \end{center}
    \caption{\sf Average test MSE throughout 10000 epochs and over 10 trials for Burgers Equation (top) and the Riccati Equation (bottom). The learning rate cycles between $10^{-2}$ and $5\cdot10^{-4}$ every $250$ training steps.}
    \label{f-CLRc}
\end{figure}

\begin{figure}[h]
    \begin{center}
         \includegraphics[scale=0.5]{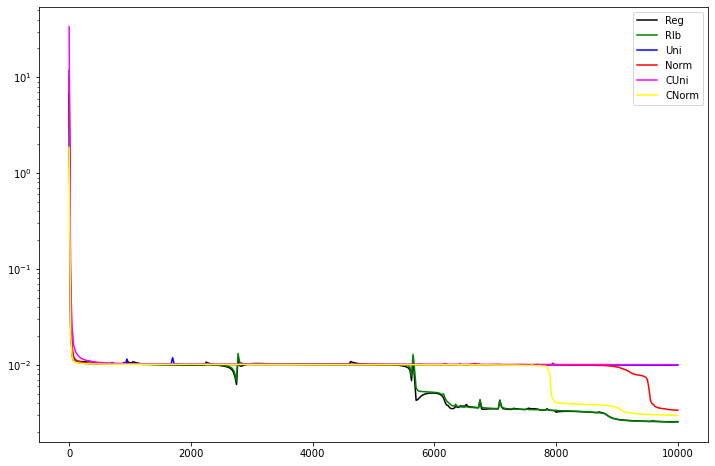}
         \includegraphics[scale=0.5]{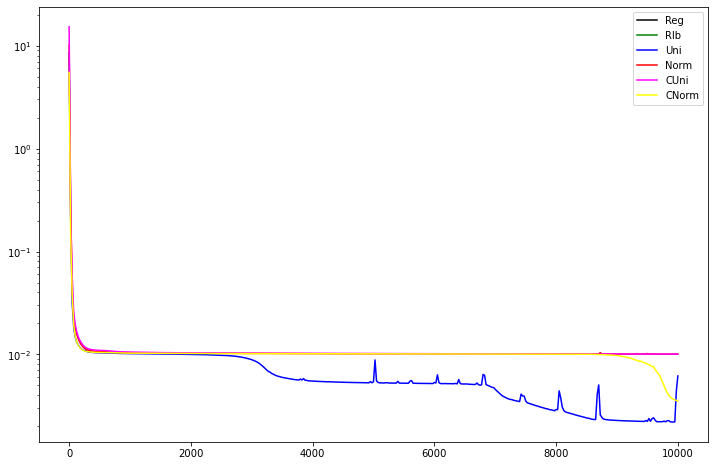}
    \end{center}
    \caption{\sf Standard loss throughout 10000 epochs for the Poisson-Boltzmann Equation with Lysozyme geometry. In the first image, the learning rate begins at $10^{-2}$ and decays to  $2\cdot10^{-3}$ and  $5\cdot10^{-4}$ after epoch $2000$ and $8000$ respectively. In the second image, the learning rate cycles between $10^{-2}$ and $5\cdot10^{-4}$ every $250$ training steps.}
    \label{f-lysL}
\end{figure}

    Now we provide a variation on the stochastic loss functions defined by Equation \ref{eq-FVRC} where we enforce that the variance approaches zero throughout the training process. In particular, suppose we have $N$ training epochs. Then at epoch $t$, we use the random variables
    \begin{align}
    \begin{split}
        &\Lambda_i \sim \text{Norm}\Big(1, 0.25 - \frac{0.25t}{N}\Big) \qquad i \in \{1,\ldots,k\}, \text{ and }\\
        &\Lambda_i \sim \text{Uni}\Big(0.5 + \frac{0.5t}{N}, 1.5-\frac{0.5t}{N}\Big) \qquad i \in \{1,\ldots,k\}.
    \end{split}
    \end{align}
    The resulting average test MSE using these random variables with a piecewise constant learning rate in given in Figure \ref{f-PLRc}. With a cyclic learning rate, the results are seen in Figure \ref{f-CLRc}. We also test these loss function on the Poisson-Boltzmann Equation in Figure \ref{f-lysL}.
    
    \subsection{Results Analysis}
\begin{figure}[h]
    \begin{center}
         \includegraphics[scale=0.5]{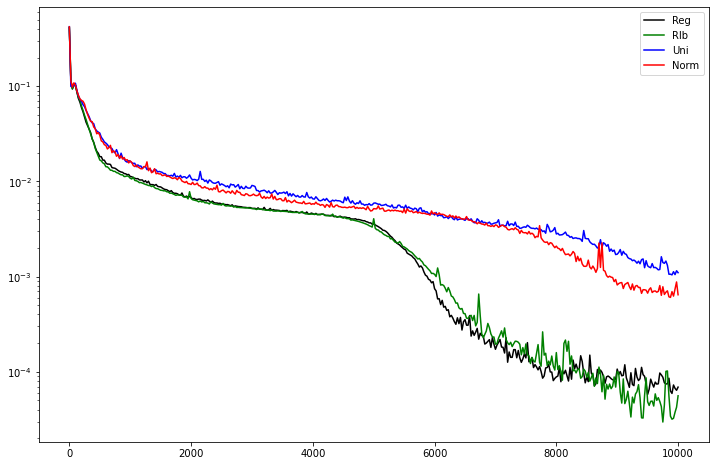}
         \includegraphics[scale=0.5]{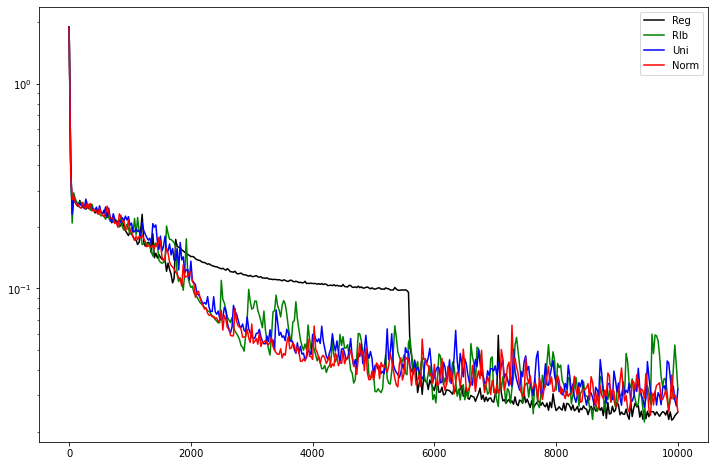}
    \end{center}
    \caption{\sf Average test MSE throughout 10000 epochs and over 10 trials for Burgers Equation (top) and the Riccati Equation (bottom). The learning rate begins at $10^{-2}$ and decays to  $2\cdot10^{-3}$ and  $5\cdot10^{-4}$ after epoch $2000$ and $8000$ respectively.}
    \label{f-PLR}
\end{figure}

\begin{figure}[h]
    \begin{center}
         \includegraphics[scale=0.5]{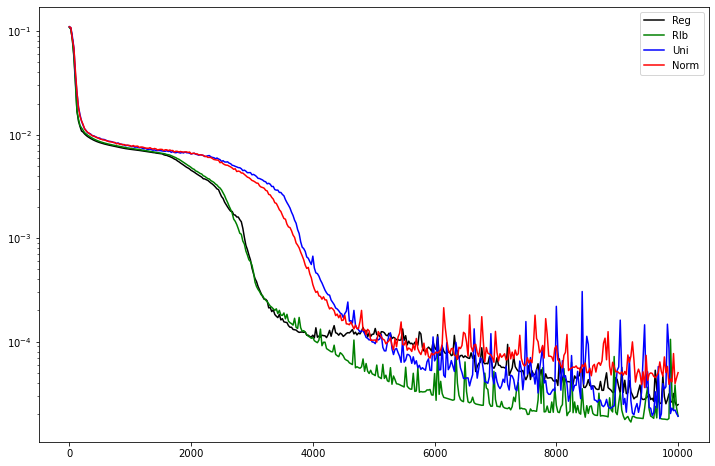}
         \includegraphics[scale=0.5]{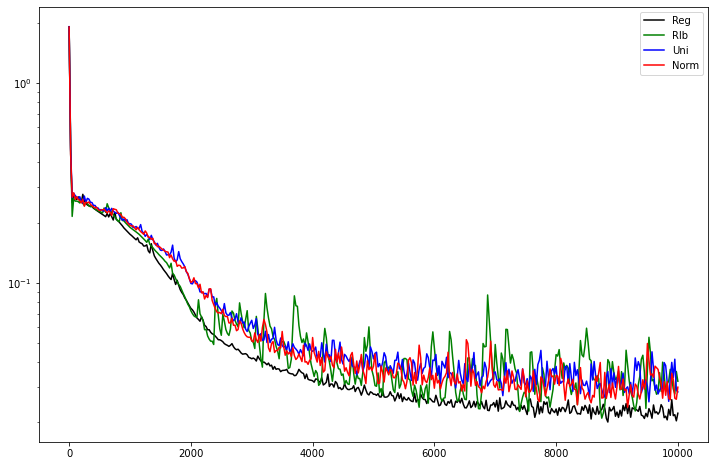}
    \end{center}
    \caption{\sf Average test MSE throughout 10000 epochs and over 10 trials for Burgers Equation (top) and the Riccati Equation (bottom). The learning rate cycles between $10^{-2}$ and $5\cdot10^{-4}$ every $250$ training steps.}
    \label{f-CLR}
\end{figure}

    Our results show that for certain PINNs, a stochastic loss function can slightly improve performance. This improvement is not consistent, and we believe that many more trials are needed before we can confidently rank the performance. One important observation is that the learning rate schedule can severely impact the performance of these loss functions, as seen in the different between the top plots in Figure \ref{f-PLR} and Figure \ref{f-CLR}. We also note that using a decreasing variance tends to outperform using a fixed variance for Burger's Equation and the Riccati equation, meanwhile using a fixed variance performed the best for the Poisson-Boltzmann equation. This could be because this problem has particularly difficult to find minima, and so the more random algorithm was more likely to stumble upon it. Finally, we observe that both uniform and normal random variables perform very similarly, with the normal random variable performing slightly better in 6 out of 10 experiments.

\section{Biological Applications}
    \subsection{Poisson-Boltzmann Equation}
    Electrostatic forces are one of the primary factors that drive molecular interactions, and the electric field generated by the charges present on molecules is rather important for biological processes \cite{rocchia_2005}. The quantification of this property has become increasingly important for fields such as drug design, molecular and cellular biology. The most popular method for structure-based drug design is molecular docking, which relies heavily on the Poisson-Boltzmann (PB) equation \cite{deruyck_brysbaert_blossey_lensink_2016}. These calculations, however, are computationally expensive and involve complex domains. In this section, we introduce the PB equation and the use of neural network algorithms to solve the equation where other numerical methods are insufficient.
     
    For a particular medium, the dielectric constant $\varepsilon$ tells us the relative electric permeability compared to a vacuum. Gauss's Law is able to relate the electrostatic potential $\phi$ in space to the electrostatic charge density $\rho$ through the classical Poisson's equation \cite{geng_2015}
    \begin{equation}
        -\nabla \cdot (\varepsilon\nabla\phi) = \rho.
    \end{equation}
    
    \begin{figure}[!t]
        \begin{center}
             \includegraphics[scale=0.235]{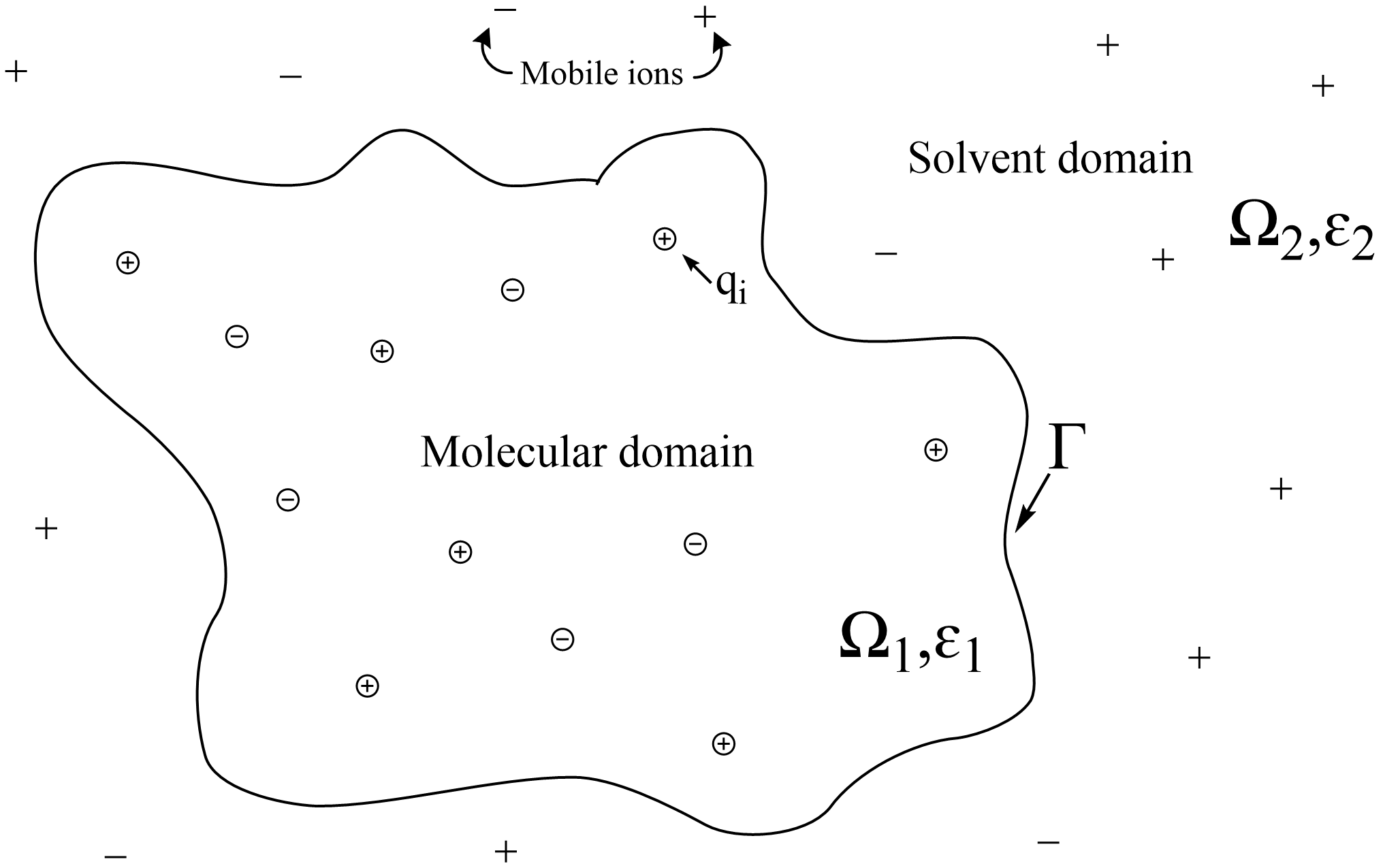}
        \end{center}
        \caption{\sf Sketch of the problem domain. Includes a complex molecular surface $\Gamma$ that contains many point charges $q_i$ inside the molecular domain $\Omega_1$, mobile ions are present in the solvent domain $\Omega_2$. The implicit model describes these mobile ions as a continuous medium with the Boltzmann distribution.}
        \label{f-PBE-D}
    \end{figure}
    
    When considering a biomolecule in a solvent, the domain is broken up into the molecular region $\Omega_1$ and the solvent region $\Omega_2$, as seen in Figure \ref{f-PBE-D}. For a particular solvent containing a biomolecule, the dielectric constant can be simplified as a piecewise constant function, with a discontinuity at the interface between the two mediums (at the interface $\Gamma$). Inside the molecular region, $\varepsilon_1 \in [1, 10]$, while in the solvent region, $\varepsilon_2 \in [70, 80]$, depending on the particular solvent. Thus, the dielectric constant can be written as 
    \begin{equation}
    \varepsilon(\mathbf{x}) =  
        \left\{
        \begin{array}{lr} 
        \varepsilon_1 & \mathbf{x} \in \Omega_1\\
        \varepsilon_2 & \mathbf{x} \in \Omega_2 \\ 
        \end{array}
        \right.
    \end{equation}
    
    Likewise, the modified Debye-Hückle parameter for the implicit solvent model is defined as a piecewise constant function discontinuous at the interface given by
    \begin{equation}
        \overline{\kappa}(\mathbf{x}) =  
        \left\{
        \begin{array}{lr} 
        0 & \mathbf{x} \in \Omega_1\\
        \sqrt{\varepsilon_2}\kappa & \mathbf{x} \in \Omega_2 \\ 
        \end{array},
        \right.
    \end{equation}
    where $\kappa$ is the Debye-Hückle screening constant, describing the attenuation in interaction casued by the presence of mobile ions in $\Omega_2$ \cite{quan_stamm_maday_2019}. This quantity relates the ionic strength of the solute $I$ to the dielectric coefficient of the medium by
    \begin{equation}
        \kappa^2 = \dfrac{8 \pi N_A I}{1000 \varepsilon k_B T},
    \end{equation}
    where $N_A$ is the Avogadro constant. Together, the Debye-Hückle constant and dielectric constant describe the solvent-solute electrostatic interactions and the accessibility of the solvent to the solute \cite{bond_chaudhry_cyr_olson_2009}.
    
    The charge density $\rho$ is represented in two parts: a summation of weighted Dirac delta functions corresponding to the point-charges of the atomic centers of the biomolecule and a Boltzmann distribution of the mobile ions \cite{geng_2015}. The charge density varies in each region, necessitating multiple equations for each domain. Finally, we must consider boundary and interface conditions. In particular, we must enforce continuity of electrostatic potential $\phi$ and flux density $\varepsilon \frac{\partial \phi}{\partial n}$ normal to the interface \cite{geng_2015}.
    
    Therefore, we can write the nonlinear Poisson-Boltzmann equation as the following:
    \begin{equation}
        -\nabla \cdot ( \varepsilon(\mathbf{x})\nabla \phi(\mathbf{x})) + \overline{\kappa}^2(\mathbf{x}) (\dfrac{k_B T}{e_c}) \sinh{\dfrac{e_c \phi (\mathbf{x})}{k_B T}} = \sum_{i=1}^{N_c} q_i \delta (\mathbf{x}-\mathbf{x_i}), \label{e-PB}
    \end{equation}
    where $\overline{\kappa}^2(x)$ is the modified Debye-Hückle parameter, $k_B$ is the Boltzmann constant, $T$ is the temperature, $e_c$ is the charge of an electron, $N_c$ is the number of atomic centers of the biomolecule centered at $\mathbf{x_i}$, $q_i$ are the corresponding charges, and $\delta$ is the Dirac delta function.

    For developing algorithms to solve the PB equation, it is necessary to break the equation up into the domains, and this can be done in the following way. The governing differential equation inside the molecular surface:
    \begin{equation}
         -\varepsilon_1 \nabla^2 \phi(\mathbf{x}) = \sum_{i=1}^{N_c}q_i\delta (\mathbf{x}-\mathbf{x_i}), \qquad \mathbf{x} \in \Omega_1. \label{e-PBi}
    \end{equation}
    The governing differential equation inside the solvent:
    \begin{equation}
        - \varepsilon_2\nabla^2\phi(\mathbf{x}) + \overline{\kappa}^2(\mathbf{x}) (\dfrac{k_B T}{e_c}) \sinh{\dfrac{e_c \phi (\mathbf{x})}{k_B T}} = 0, \qquad \mathbf{x} \in \Omega_2.
        \label{e-PBo}
    \end{equation}
    Continuity of electrostatic potential $\phi$ and flux density $\varepsilon \frac{\partial \phi}{\partial n}$ normal to the dielectric boundary $\Gamma$:
    \begin{equation}
        \phi_1(\mathbf{x}) = \phi_2(\mathbf{x}), \qquad \varepsilon_1 \frac{\partial \phi_1(\mathbf{x})}{\partial n} = \varepsilon_2 \frac{\partial \phi_2(\mathbf{x})}{\partial n}, \qquad \mathbf{x} \in \Gamma.
    \end{equation}
    The boundary condition:
    \begin{equation}
        \lim_{|\mathbf{x}|\to\infty} \phi (\mathbf{x}) = 0. \label{e-BC}
    \end{equation}
    
    \subsection{Challenges}
    The Poisson-Boltzmann equation presents many challenges, especially with numerical solutions. The nonlinear portion of the equation presents many issues for classical methods and complicates the solving process of the equation.
    
    Classical methods, such as the finite element method, require fine mesh to represent the biomolecule. The molecular surface $\Gamma$ is complex, as such, many proteins require high resolution to represent the surface with acceptable accuracy, making calculations computationally expensive.
    
    The atomic centers of the biomolecule are represented by singular point charges through the Dirac delta function, as seen in equation (\ref{e-PBi}). The discontinuous source terms lead to numerical instability.
    
    Furthermore, the dielectric constant is defined as a discontinuous function across the molecular surface. This leads to the discontinuity of the electric field $-\nabla \phi$ as well as the non-smoothness of the electrostatic potential $\phi$ across the interface $\Gamma$ \cite{geng_2015}.
    Finally, the domain is unbounded, as imposed by the infinite boundary condition in (\ref{e-BC}).
    
    \subsection{Algorithm Formulation}
    In order to deal with the many challenges of solving the nonlinear Poisson-Boltzmann equation, we propose some modifications to the governing differential equations. Similar to previous results in the paper, a PINN approach was utilized for the basic neural network architecture to solve for the electrostatic potential $\phi$ \cite{https://doi.org/10.48550/arxiv.1711.10561}. Due to the nature of neural networks, the nonlinearity and complex domain of the PB equation do not present as big of a challenge when compared with other numerical methods. Thus, our focus was on improving the algorithms for dealing with other challenges.
    
    Singular point charges represented by a summation of Dirac delta function can be approximated as a continuous probability density function to eliminate the singularity \cite{Huang}. In our application, we used a Gaussian distribution to represent the point charges as
    \begin{equation}
      \delta (\mathbf{x}-\mathbf{x_i}) \approx  \frac{1}{\sigma \sqrt{2 \pi}} {e}^{-\frac{( \mathbf{x}-\mathbf{x_i})^2}{2\sigma^2}}
    \end{equation}
    with a small value of $\sigma$, thereby increasing numerical stability and decreasing computational cost, as the source charges are no longer modeled by a singularity.
    
    To increase computational efficiency, the infinite boundary conditions can be truncated at a finite distance from the biomolecular centers \cite{bond_chaudhry_cyr_olson_2009}. As suggested by \cite{bond_chaudhry_cyr_olson_2009}, the new boundary conditions can be described as a linear combination of Helmholtz Green's functions, given by
    \begin{equation}
        \phi (\mathbf{x}) = \frac{e_c}{k_B T} \sum_{i=1}^{N_c} \frac{q_i}{\varepsilon_s |\mathbf{x} - \mathbf{x_i}|} \exp\left(\frac{-\overline{\kappa} | \mathbf{x} - \mathbf{x_i} |}{\sqrt{\varepsilon_s}}\right). \label{e-green}
    \end{equation}
    
    As discussed earlier in the paper, the DCSNN technique described in \cite{https://doi.org/10.48550/arxiv.2106.05587} is utilized to deal with the discontinuous constant coefficients across the two domains. Using the protocol described in equations \ref{e-DCSNN1} and \ref{e-DCSNN2}, an extra feature is introduced to extend the discontinuous function in $d$ dimensions to a continuous function in $d+1$ dimensions. We use Equation \ref{e-PBi} to defined $f^+(\mathbf{x})$ and Equation \ref{e-PBo} to define $f^-(\mathbf{x})$. Thus our equations define a continuous solution of the form given in Equation \ref{e-DCSNN2}, which can in theory be approximated with a neural network by Theorem \ref{t-UAT}. Using these modifications, we move on to the next section to solve the nonlinear PB equation.
    
    \subsection{Lysozyme}
    
\begin{figure}[!b]
        \begin{center}
             \includegraphics[scale=0.2]{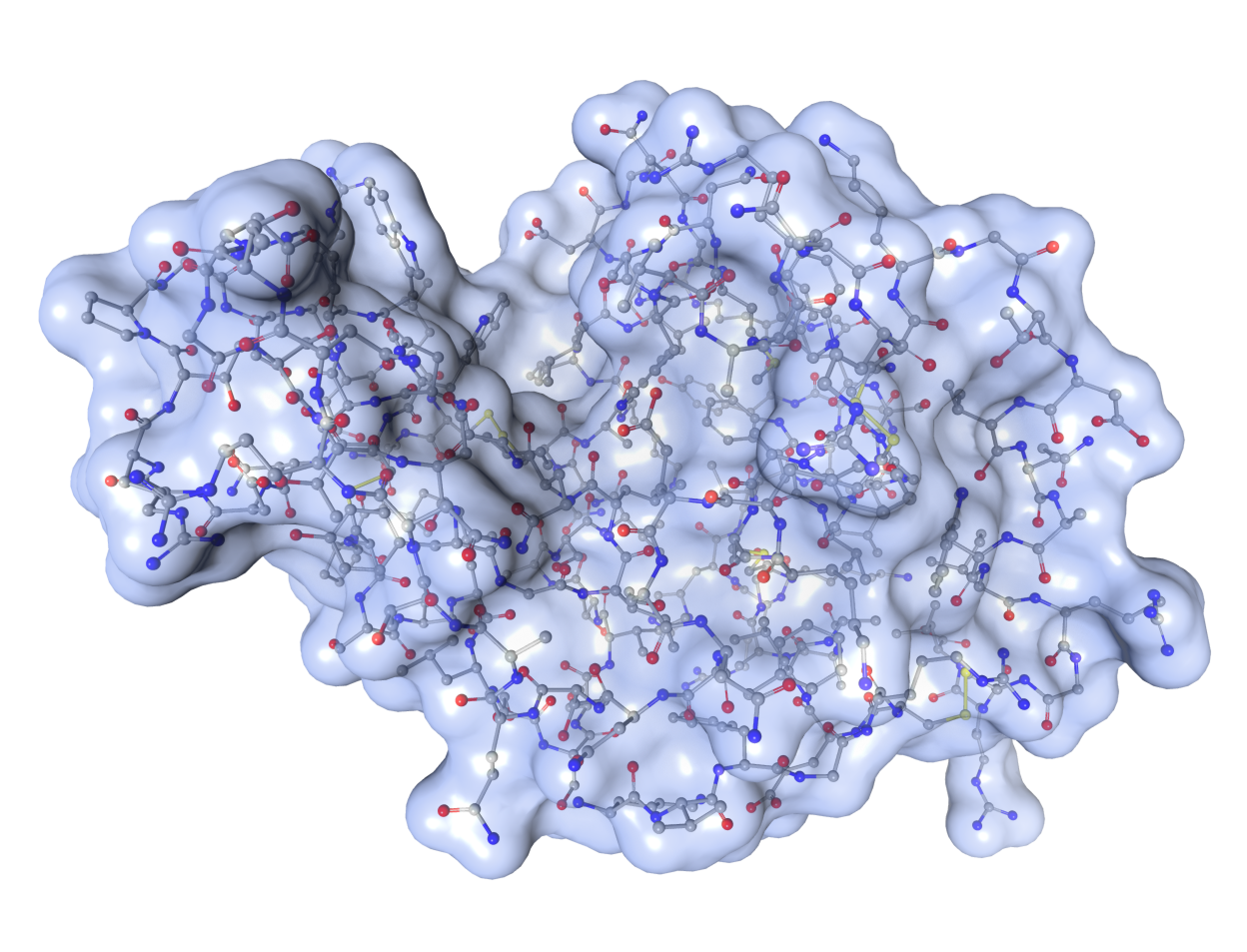}
        \end{center}
        \caption{\sf Molecular model of Lysozyme protein (1HEL)\cite{wilson_malcolm_matthews_1993}. The molecular surface (blue) represents the interface for our model. Rendered using software from  \cite{tomasello_armenia_molla_2020}. }
        \label{f-1HEL}
    \end{figure}

    We finish this section by considering a model of particular interest, the enzyme Lysozyme. This model is commonly used as a test case for numerical methods for the Poisson-Boltzmann equation. Using chicken egg white Lysozyme (1HEL from the PDB) \cite{wilson_malcolm_matthews_1993}, we create our model. As seen in Figure \ref{f-1HEL}, a molecular surface is constructed by a space-filling model using the associated van der Waals radii for each atomic center. The point charges $q_i$ are calculated and assigned to each atomic center. For simplicity, structure data from \cite{Cooper2013} is utilized as to aid in comparing results.

\begin{figure}[!h]
    \centering
    \begin{subfigure}{.5\textwidth}
      \centering
      \includegraphics[width=\linewidth]{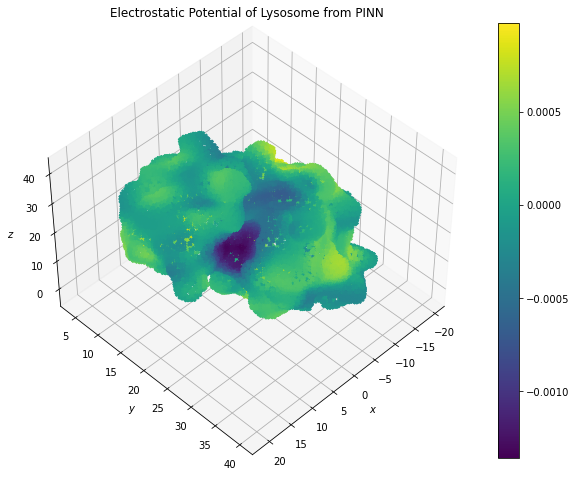}
      \caption{Electrostatic potential on Lysozyme surface using PINN \& PyGBe Data}
      \label{fig:sub1}
    \end{subfigure}%
    \begin{subfigure}{.5\textwidth}
      \centering
      \includegraphics[width=\linewidth]{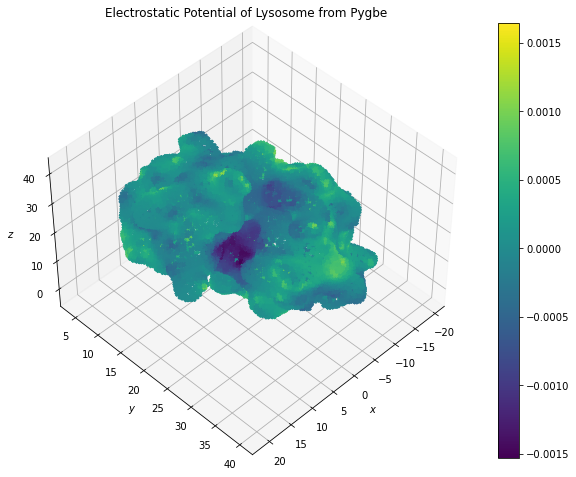}
      \caption{Electrostatic potential on Lysozyme surface from PyGBe}
      \label{fig:sub2}
    \end{subfigure}
    \caption{Comparing numerical solution to non-linear Poisson Boltzmann using PINN aided by PyGBe data (left) to numerical solution to linearized Poisson Boltzmann equation using PyGBe (right)}
    \label{f-LysM}
    \end{figure}

    Using real world constants in our model, we were unable to sufficiently train the PINN. In order to improve training, we incorporated data from PyGBe \cite{Cooper2013}, a numerical solver for the linearized Poisson-Boltzmann. It also seems that our boundary conditions were inconsistent with the those used by PyGBe as that model also included a stern layer. Replacing our boundary condition term with a PyGBe data term, we were able to train a neural network to approximate the PyGBe data while also minimizing the loss due to the non-linear Poisson Boltzmann equation. This results in the model seen in Figure \ref{f-LysM}. In order to quantify the accuracy of our model, we need a numerical solution to the non-linear Poisson-Boltzmann equation on Lysozyme, which we could not obtain.
    
\section{Discussion and Future Work}
We introduced several modifications of the standard loss functions used in PINNs. In particular, we explored variations which used different norms, as well as a novel stochastic scaling approach. As previously noted, we have shown that a stochastic loss function can slightly improve testing performance for certain PINNs - allowing for more accurate neural network based modeling of phenomena governed by differential equations. Although we only show minor improvements, greater benefits may be possible from more carefully tuned hyperparameters, or more sophisticated stochastic scalings. In particular, we have seen that learning rate can greatly effect the performance of stochastic loss functions, and a thorough investigation of this relationship is needed. Moreover, we only considered stochastic coefficients with constant expectations. One could consider an adaptive expectation which follows an update rule such as SoftAdapt. 

We also show that these methods could reach a lower loss in a PINN designed to model the Lysozyme protein. This provides evidence that stochastic loss functions could be particularly effective in solving PDEs with complex solutions defined by many conditions. Regardless of this lower loss, the PINN alone could not produce an accurate model. Using a data term from other models of Lysozyme, we were able to produce a convincing model, which in theory approximates a more accurate differential equation than the data that was used. Further experiments are needed to verify the accuracy of this model.



\section*{Acknowledgements}
The authors would like to thank the faculty of the UNC Charlotte Math Department and fellow students at the Math Research at UNC Charlotte summer 2022 REU for their continued support and encouragement throughout this program. They would also like to acknowledge their mentor, Dr. Duan Chen, for his guidance throughout this project. This project has been funded by and made possible by the National Science Foundation under the NSF-REU program grant no. DMS-2150179.

\printbibliography

\end{document}